\documentclass{article}

\usepackage{microtype}
\usepackage{graphicx}
\usepackage{booktabs}

\usepackage[draft]{hyperref}

\usepackage{times,color}
\usepackage{listings}
\usepackage{amssymb}
\usepackage{dsfont}
\usepackage{paralist}
\usepackage{graphicx}
\usepackage{wrapfig}
\usepackage{listings}
\usepackage{subcaption}
\usepackage{placeins}
\usepackage{float}
\usepackage{booktabs}
\usepackage{hyperref}
\usepackage[algo2e]{algorithm2e}
\usepackage[font=small,labelfont=bf]{caption}
\usepackage[capitalise]{cleveref}
\usepackage{enumitem}

\usepackage{ifthen}

\newcommand{\DONE}[1]{}

\newcommand{\ignore}[1]{}

\newcounter{programlinenumber}

\newboolean{TR}
\setboolean{TR}{false}
\ifthenelse{\boolean{TR}}{

\newcommand{\TrOnly}[1]{#1}
\newcommand{\SubOnly}[1]{}
\newcommand{\TrOnlyInFootnote}[1]{#1}
\newcommand{\TrOnlyInTable}[1]{#1}}
{

\newcommand{\TrOnly}[1]{}
\newcommand{\SubOnly}[1]{#1}
\newcommand{\TrOnlyInFootnote}[1]{}
\newcommand{\TrOnlyInTable}[1]{}}

\newcommand{\lstar}{L$^*$ }
\newcommand{\lstarnogap}{L$^*$}

\newtheorem{property}{Property}

\newcommand{\hypoaut}{\mathcal{A}}
\newcommand{\fhypoaut}{f_{\hypoaut}}

\newcommand{\RETURN}{\STATE \textbf{return }}

\newcommand{\hiddentext}[1]{}

\newcommand{\para}[1]{\vspace{3pt}\noindent\textbf{\textit{#1}}}

\renewcommand{\phi}{\varphi}

\usepackage{color}
\definecolor{lightgray}{rgb}{.9,.9,.9}
\definecolor{darkgray}{rgb}{.4,.4,.4}
\definecolor{purple}{rgb}{0.65, 0.12, 0.82}
\lstdefinelanguage{JavaScript}{
  keywords={break, case, catch, continue, debugger, default, delete, do, else, false, finally, for, function, if, in, instanceof, new, null, return, switch, this, throw, true, try, typeof, var, void, while, with},
  morecomment=[l]{//},
  morecomment=[s]{/*}{*/},
  morestring=[b]',
  morestring=[b]",
  ndkeywords={class, export, boolean, throw, implements, import, this},
  keywordstyle=\color{blue}\bfseries,
  ndkeywordstyle=\color{darkgray}\bfseries,
  identifierstyle=\color{black},
  commentstyle=\color{purple}\ttfamily,
  stringstyle=\color{red}\ttfamily,
  sensitive=true
}

\usepackage[accepted]{icml2018}

\icmltitlerunning{Extracting Automata from Recurrent Neural Networks Using Queries and Counterexamples}

\begin{document}

\twocolumn[
\icmltitle{Extracting Automata from Recurrent Neural Networks\\
			 Using Queries and Counterexamples}

\icmlsetsymbol{equal}{*}

\begin{icmlauthorlist}
\icmlauthor{Gail Weiss}{tech}
\icmlauthor{Yoav Goldberg}{bar}
\icmlauthor{Eran Yahav}{tech}
\end{icmlauthorlist}

\icmlaffiliation{tech}{Technion, Haifa, Israel}
\icmlaffiliation{bar}{Bar Ilan University, Ramat Gan, Israel}

\icmlcorrespondingauthor{Gail Weiss}{sgailw@cs.technion.ac.il}

\icmlkeywords{Recurrent Neural Networks, Automata, Exact Learning}

\vskip 0.3in
]

\printAffiliationsAndNotice{}

\begin{abstract}
We present a novel algorithm that uses exact learning and abstraction to extract a deterministic finite automaton describing the state dynamics of a given trained RNN. 
We do this using Angluin's \lstar algorithm as a learner and the trained RNN as an oracle. 
Our technique efficiently extracts accurate automata from trained RNNs, 
even when the state vectors are large and require fine differentiation.
\end{abstract}

\section{Introduction}\label{Se:Intro}

Recurrent Neural Networks (RNNs) are 
a class of neural networks used to 
process sequences of arbitrary lengths. 
An RNN receives an input sequence timestep by timestep, 
returning a new \emph{state vector} after each step. 
For classification tasks, this is followed by passing the state vectors to a multi-class classification component, 
which is trained alongside the RNN and
returns a classification for the sequence.
We call a combination of an RNN and a binary classification component an \emph{RNN-acceptor}. 

RNNs are central to deep learning, 
and natural language processing in particular. 
However, while they have been shown to reasonably approximate a variety of languages, 
what they eventually learn is unclear. 
Indeed, several lines of work attempt to 
extract clear rules for their decisions
~\cite{NNExtractionOverview,NNExtraction,NNExtractionFuzzyClustering}.

\para{Motivation} Given an RNN-acceptor $R$ trained 
over a finite alphabet $\Sigma$, 
our goal is to extract a deterministic finite-state automaton (DFA) $A$ that 
classifies sequences in a manner
observably equivalent to $R$. 
(Ideally, 
we would like to obtain a DFA that accepts \emph{exactly} 
the same language as the network,
but this is a much more difficult task.)

We approach this task using \emph{exact learning}. 

\para{Exact Learning} In the field of exact learning, \emph{concepts} (sets of instances) can be learned precisely from a \emph{minimally adequate teacher}---an oracle capable of answering two query types \cite{TeachingComplexity}:
\begin{itemize}
	\setlength{\itemsep}{0pt}
	\item \emph{membership queries}: label a given instance
	\item \emph{equivalence queries}: state whether a given hypothesis (set of instances) is equal to the concept held by the teacher. If not, return an instance on which the hypothesis and the concept disagree (a \emph{counterexample}).
\end{itemize}
The \lstar algorithm \cite{Lstar} is an exact learning algorithm for learning a DFA from a minimally adequate teacher 
for some
regular language $L$. 
In this context, 
the concept is $L$, 
and the instances are words over its alphabet.

We designate a trained RNN\footnote{In 
what follows, when understood from context, 
we use the term RNN to mean RNN-acceptor.}
as teacher for the \lstar algorithm,
in order to extract a DFA representing its behavior. 
The RNN is used trivially to answer membership queries: input sequences are fed to the network for classification.
The main challenge 
in this setting
is answering equivalence queries.

\para{Problem Definition: Equivalence Query} 
Given an RNN-acceptor $R$ trained 
over a finite alphabet $\Sigma$,
and a DFA $\hypoaut$ over $\Sigma$, 
determine whether $R$ and $\hypoaut$ are equivalent, 
and return a counterexample $w\in\Sigma^*$ if not.

As this problem is likely to be intractable, 
we use an approximation.
One approach would be random sampling;
however, 
should $R$ and $\hypoaut$ be similar,
this may take time.

\para{Our Approach} We use \emph{finite abstraction} of the RNN $R$ to answer equivalence queries. The finite abstraction and the \lstar DFA $\hypoaut$ act as two hypotheses for the RNN ground truth, and must at least be equivalent to each other in order to be equivalent to $R$.
Whenever the two disagree on a sample,
we find its true classification in $R$, 
obtaining through this either a counterexample to $\hypoaut$ or 
a refinement to the abstraction.

Our approach is guaranteed never to return an incorrect counterexample 
nor invoke an unnecessary refinement;
i.e., \emph{it yields no false negatives}.
As far as we know,
this is the first attempt to apply exact learning to a given RNN.

\para{Main Contributions} 
\begin{itemize}[topsep=0pt]
\item We present a novel and general framework for extracting automata from trained RNNs. We use the RNNs as teachers in an exact learning setting.
\item We implement\footnote{\texttt{www.github.com/tech-srl/lstar\_extraction}} the technique and show its ability to 
extract descriptive automata in settings where 
previous approaches fail. 
We demonstrate its effectiveness on 
modern RNN architectures. 
\item We apply our technique to RNNs trained to $100\%$ train and test accuracy on simple languages, and 
discover in doing so that 
some RNNs have not 
generalized to the intended concept.
Our method easily reveals and produces 
\emph{adversarial inputs}---words 
misclassified by the trained RNN and not present in the train or test set.
\end{itemize}

\section{Related Work}\label{Se:relatedWork}
DFA extraction from RNNs was extensively explored by Giles and colleagues; see Wang et al. (\citeyear{NNExtractionGiles2017}) and Jacobsson (\citeyear{NNExtractionOverview}) for a partial survey.

Broadly, the approaches work by defining a finite partitioning
of the real-valued RNN state space 
and then exploring the network transitions in the partitioned space, 
using techniques such as BFS exploration \cite{NNExtraction}
and other transition-sampling approaches. 
The approaches differ mainly in their choice and definition of partitioning.  

These works generally use second order RNNs~\cite{high_order_nets}, which are shown to better map DFAs than first-order Elman RNNs~\cite{Elman,first_second_order,ExtractionComparisonFirstSecondOrders}. In this work, however, we will focus on GRUs~\cite{GRU1,GRU2} and LSTMs~\cite{LSTM}, 
as they are more widely used in practice.

One approach to state space partitioning is
to divide each dimension into
$q$ equal intervals, with $q$ being 
the \emph{quantization level} \cite{NNExtraction}.
This approach suffers from inherent state space explosion
and does not scale to the networks used in practice today.
(The original paper demonstrates the technique on 
networks with 8 hidden values, 
whereas today's can have hundreds to thousands).

Another approach is to fit an unsupervised classifier 
such as k-means 
to a large sample set of reachable network states \cite{NNExtractionFuzzyClustering,NNExtractionClustering93}.
The number of clusters $k$ generated with these classifiers
is a parameter
that might greatly affect extraction results,
especially if it is too small.
The sample states can be found 
by a simple BFS exploration of the network state space to a certain depth,
or 
by recording all state vectors reached by the network
when applied to its train set (if available).

An inherent weakness of both these approaches is that 
the partitioning is set before the extraction begins,
with no mechanism for recognizing and overcoming
overly coarse behavior. 
Both methods thus face the challenge of choosing 
the best parameter value for extraction.
They are generally applied 
several times with different parameter values, 
after which the `best' 
DFA is 
chosen according to a heuristic.

Current techniques treat all the dimensions of an RNN as a single state. In future work, it may be interesting to make the distinction between `fast' and `slow' internal states as introduced in the differential state framework unifying GRUs and LSTMs~\cite{slow-fast}.

\section{Background}\label{Se:Background}
\para{Recurrent Neural Networks and RNN acceptors} 
An RNN is a parameterized function $g_R(h,x)$ that 
takes as input a state-vector $h_t\in\mathbb{R}^{d_s}$ and an 
input vector $x_{t+1}\in\mathbb{R}^{d_i}$ and returns 
a state-vector $h_{t+1}\in\mathbb{R}^{d_s}$. 
An RNN can be applied to a sequence $x_1,...,x_n$ by 
recursive application of the function $g_R$ to the vectors $x_i$.
To use a set of discrete symbols as an input alphabet, 
each symbol is deterministically mapped to an input vector using 
either a one-hot encoding or an embedding matrix. 
As we are only interested in the internal network transitions, 
we use one-hot encoding in this work.
For convenience, we refer to input symbols and 
their corresponding input vectors interchangeably.
We denote the state space of a network $R$ by 
$S_R=\mathbb{R}^{d_s}$.
For multi-layered RNNs, 
where several layers each have their own state vector, 
we consider the concatenation of these vectors as the state vector of the entire network.
In a binary \emph{RNN-acceptor}, 
there is an additional function 
$f_R: S_R \rightarrow\{Acc,Rej\}$ 
that classifies the RNN's state vectors.
An RNN-acceptor $R$ is defined by the pair of functions $g_R,f_R$.

\para{Network Abstraction}
Given a neural network $R$ with state space $S$ and alphabet $\Sigma$, 
and a partitioning function $p \colon S\rightarrow\mathbb{N}$, 
Omlin and Giles (\citeyear{NNExtraction})
presented a method for 
extracting
a DFA for which every state is a partition from $p$,
and the state transitions and classifications are defined by
a single
sample from each partition. 
The method is effectively a BFS exploration of 
the partitions defined by $p$, 
beginning with $p(h_0)$,
where $h_0$ is the network's initial state,
and continuing according to the network's transition function $g_R$.

We denote by $A^{R,p}$ the DFA extracted by this method from a network $R$ and partitioning $p$, and denote all its related sets and functions by subscript $R,p$.

\para{The \lstar Algorithm}
The \lstar algorithm is an exact learning algorithm for 
extracting a DFA from any teacher that can answer 
\emph{membership queries} (label a given word) 
and \emph{equivalence queries} (accept or reject a given DFA, with a counterexample if rejecting).
We know that L$^*$ always proposes a minimal
DFA in equivalence queries
and utilize this in our work.
Beyond this, we treat the algorithm as a black box.
A short review is provided in the supplementary material.

\section{Learning Automata from RNNs using L*}\label{Se:RNNTeacher}

We build an RNN-based teacher for \lstar as follows:

\para{For membership queries}, we use the RNN classifier directly,
checking whether it accepts or rejects the given word. 

\para{For equivalence queries:} Given a proposed DFA $\hypoaut$, 
we compare it to 
abstractions $A^{R,p}$ of the network $R$,
beginning with some initial partitioning 
$p$ of $S_R$.
If we find a disagreement between $\hypoaut$ and an abstraction $A^{R,p}$,
we use $R$
to determine whether to return it as a counterexample 
or to refine $p$ and restart the comparison.

In theory this continues until $\hypoaut$ and $A^{R,p}$ converge, i.e., 
are equivalent.
In practice, for some RNNs this may take a long time 
and yield a large DFA ($>$30,000 states). 
To counter this, we place time or size limits on the interaction,
after which the last \lstar DFA, $\hypoaut$,
is returned. 
We see that these DFAs still 
generalize well to their respective networks.\footnote{We could also return the last abstraction, $A^{R,p}$, 
and focus on refining $p$ over returning counterexamples.
But the abstractions are often less accurate. 
We suspect this is due to 
the lack of `foresight' $A^{R,p}$ has in comparison to
\lstarnogap's many separating suffix strings.}

\para{Note} Convergence of $A^{R,p}$ and $\hypoaut$ does
not guarantee that 
$R$ and $\hypoaut$ are equivalent. 
Providing such a guarantee would be an interesting direction 
for future work.

\section{Notations}\label{Se:Notations}

\para{Automaton and Classification Function} 
For a deterministic automaton 
$A =\langle \Sigma, Q, q_0, F, \delta \rangle$, 
$\Sigma$ is its alphabet, 
$Q$ the set of automaton states, 
$F \subseteq Q$ the set of accepting states, 
$q_0 \in Q$ the initial state, 
and $\delta : Q \times \Sigma \rightarrow Q$ its transition function. 
We denote by $\hat{\delta}:Q\times \Sigma^* \rightarrow Q$ 
the recursive application of $\delta$ to a sequence, 
i.e., for every $q\in Q$, $\hat{ \delta }(q,\varepsilon)=q$, 
and for every $w\in\Sigma^*$ and $\sigma\in\Sigma$, 
$\hat{\delta}(q,w\cdot\sigma)=\delta(\hat{\delta}(q,w),\sigma)$. 
For convenience, 
we add the notation $f:Q\rightarrow\{Acc,Rej\}$ as 
the function giving the classification of each state, 
i.e., $f(q)=Acc \iff q\in F$.

\para{Binary RNN-acceptor}
For a binary RNN-acceptor, 
we denote by $h_{0,R}$ the initial state of the network, 
and by $\hat{g_R}:S_R\times \Sigma^*\rightarrow S_R$ 
the recursive application of $g_R$ to a sequence, 
i.e., for every $h\in S_R$, $\hat{g_R}(h,\varepsilon)=h$, 
and for every $w\in\Sigma^*$ and $\sigma\in\Sigma$, $\hat{g_R}(h,w\cdot\sigma)=g_R(\hat{g_R}(h,w),\sigma)$. 
We drop the subscript $R$ when it is clear from context.

We note that a given RNN-acceptor can be interpreted as a deterministic, though
possibly infinite, state machine.

\para{Shorthand} 
As an abuse of notation, for any DFA or RNN classifier $C$ with state transition function $t_C$, state classification function $f_C$, and initial state $q_{C,0}$,
we use 
$\hat{t_C}(w)$ to denote $\hat{t_C}(q_{C,0},w)$,  
$f_C(q,w)$ to denote $f_C(\hat{t_C}(q,w))$,
and $f_C(w)$ to denote $f_C(\hat{t_C}(q_{C,0},w))$. 
Within this notation, 
the classifications of a word $w\in\Sigma^*$ by an automaton $A$
and a binary RNN-acceptor $R$ with respective classification functions $f_A$ and $f_R$ are given by $f_A(w)$ and $f_R(w)$.

\section{Answering Equivalence Queries}\label{Se:EquivalenceQueries}
\subsection{Overview}
Given a network $R$, 
a partitioning function $p:S\rightarrow\mathbb{N}$ over its state space $S$, 
and a proposed minimal automaton $\hypoaut$, 
we wish to check whether $R$ is equivalent to $\hypoaut$, 
preferably exploring as little of $R$'s behavior as necessary to respond. 

We search for 
a disagreeing example $w$ between $\hypoaut$ 
and the abstraction $A^{R,p}$,
by parallel traversal of the two. 
If one is found, 
we check its true classification in $R$.
If this disagrees with $\hypoaut$, 
$w$ is returned as a counterexample;
otherwise,
$p$ is refined (\cref{Se:Quantisation}) and the traversal 
begins again.\footnote{If
the refinement does not affect 
any states traversed so far,
this is equivalent to 
fixing the current state's abstraction and continuing.}

Every counterexample $w$ returned by our method is inherently true, 
i.e., satisfies $f_\hypoaut (w) \neq f_R(w)$. 
From this and the minimality 
of \lstar equivalence queries, 
we obtain:

\begin{property} Every separate state in the final extracted automaton $\hypoaut$ is justified by concrete input to the network. \label{Thm:Lstar} \end{property} 

In other words, all complexity in a DFA extracted from a given RNN $R$ is a result of the inherent complexity of $R$. 
This is in contrast to other methods, 
in which incorrect partitioning of the network state space may lead to unnecessary complexity in the extracted DFA,
even after minimization.
Moreover, 
our method refines the partitioning 
only when it is proven too coarse to correctly represent the network:

\begin{property} Every refinement to the partitioning function $p:S\rightarrow\mathbb{N}$ is justified by concrete input to the network. \label{Thm:Abs} \end{property}

This is important, 
as the search for counterexamples runs atop an extraction of the abstraction $A^{R,p}$, 
and so unnecessary refinements---which 
may lead to state space explosion---can 
make the search so slow as to be impractical.

For clarity, we henceforth refer to the continuous network states $h\in S$ as R-states, the abstracted states in $A^{R,p}$ as A-states, and the states of the \lstar DFAs as L-states.

\subsection{Parallel Exploration}

The key intuition to our approach is that $\hypoaut$ is minimal, 
and so each A-state should---if 
the two DFAs are equivalent---be 
equivalent to exactly one L-state, 
w.r.t. classification and 
projection of transition functions. 
The extraction of $A^{R,p}$ 
is effectively a BFS traversal of $A^{R,p}$, 
allowing us to associate between states in the 
two DFAs 
during its extraction.

We refer to bad associations, 
in which an accepting A-state is associated with a rejecting L-state or vice versa, 
as \emph{abstract classification conflicts}, 
and to multiple but disagreeing associations, 
in which one A-state is associated with two different 
L-states, 
as \emph{clustering conflicts}. (The inverse case, 
in which one 
L-state is associated with several A-states, 
is not necessarily a problem, 
as $A^{R,p}$ is not necessarily minimal.)

We may also assert 
that the classification of 
each R-state $h$ encountered while extracting $A^{R,p}$ 
is identical to that of the L-state $q_{\hypoaut}\in Q_{\hypoaut}$ that the parallel traversal of $\hypoaut$ reaches during the exploration. 
As the classification of an A-state is determined 
by the R-state with which it was first reached, 
this also covers all abstract classification conflicts. 
We refer to failures of this assertion as \emph{classification conflicts}, and check only for them and for clustering conflicts.

\subsection{Conflict Resolution and Counterexample Generation} \label{Se:Resolution}
We assume an initial 
partitioning $p:S\rightarrow \mathbb{N}$ 
of the R-state space 
and a refinement operation $\mathit{ref}:p,h,H\mapsto p'$ which receives a partitioning $p$, an R-state $h$, and a set of R-states $H\subseteq S\setminus \{h\}$, and returns a new partitioning $p'$ satisfying:
\begin{enumerate}
	\setlength{\itemsep}{0pt}
	\item $\forall h_1\in H$, $p'(h)\neq p'(h_1)$, and
	\item $\forall h_1,h_2\in S$, $p(h_1)\neq p(h_2)\Rightarrow p'(h_1)\neq p'(h_2)$.
\end{enumerate}
(In practice, condition 1 may be relaxed to separating at least one of the vectors in $H$ from $h$, and our method can and has overcome imperfect splits.) 

\para{Classification conflicts} occur when some $w\in\Sigma^*$ 
for which $f_R(w)\neq \fhypoaut(w)$ 
has been traversed during $A^{R,p}$'s extraction. 
We resolve them by returning $w$ as a counterexample.

\para{Clustering conflicts} occur when the parallel exploration associates an A-state $q\in Q_{R,p}$
with an L-state $q_2$,
after $q$ has already been associated with an L-state $q_1\neq q_2$. 
As $\hypoaut$ is minimal, 
$q_1$ and $q_2$ cannot be equivalent.
It follows that if $w_1,w_2\in\Sigma^*$ are the BFS traversal paths
through which $q$ was associated with $q_1,q_2\in Q_{\hypoaut}$, 
then there exists some
differentiating 
sequence $s\in\Sigma^*$ 
for which $f_{\hypoaut}(q_1,s)\neq f_{\hypoaut}(q_2,s)$, 
i.e.,
for which 
$f_{\hypoaut}(w_1{\cdot}s)\neq f_{\hypoaut}(w_2{\cdot} s)$. 
Conversely, 
the arrival of $w_1$ and $w_2$ at the same A-state $q\in A^{R,p}$ 
gives
 $f_{R,p}(w_1{\cdot} s)=f_{R,p}(q,s)=f_{R,p}(w_2{\cdot} s)$.

It follows that $\hypoaut$ and $A^{R,p}$ disagree on the classification of either $w_1{\cdot} s$ or $w_2{\cdot} s$, 
and so 
necessarily 
at least one is not equivalent to $R$. 
We pass 
$w_1{\cdot} s$ and $w_2{\cdot} s$ 
through $R$ for their true classifications. 
If $\hypoaut$ is at fault, 
the sequence on which $\hypoaut$ and $R$ disagree 
is returned as a counterexample.
Otherwise, necessarily, $f_R(w_1{\cdot} s)\neq f_R(w_2{\cdot} s)$,
and so $A^{R,p}$ should satisfy 
$\hat{ \delta_{R,p} }(w_1)\neq\hat{ \delta_{R,p} }(w_2)$. 
The R-states $h_1=\hat{g}(w_1)$ and $h_2=\hat{g}(w_2)$
are passed, 
along with $p$, 
to $\mathit{ref}$, 
to yield a new, finer, partitioning $p'$  for which 
$\hat{ \delta_{R,p'} }(w_1)\neq\hat{ \delta_{R,p'} }(w_2)$.

This reasoning applies to
$w_2$ with \emph{all} paths $w'$ 
that have reached 
$q$ without conflict before $w_2$. 
As such, 
the classifications of \emph{all} words $w'{\cdot} s$ are tested against $R$, 
prioritizing 
returning a counterexample over refining $p$.
If a refinement is triggered, 
then $h=\hat{g}(w_2)$ is split from the set of
R-states $h'=\hat{g}(w')$.

\cref{Equivalence} shows pseudocode for this equivalence checking. 
In it,
all mappings except one are unique and defined before they are accessed.
The exception is \texttt{Paths},  
as we might reach the same R-state $h\in S_R$ more than once, 
by different paths. 
This can be remedied by maintaining in \texttt{Paths} not single paths but lists of paths.
   
Our experiments showed that long counterexamples often caused $\hypoaut$ to blow up, 
without generalizing well. 
Thus, we always return the shortest available counterexample. 

\begin{algorithm}
   \caption{Pseudo-code for equivalence checking of an RNN $R$ 
   and minimal DFA $\hypoaut$, with initial 
   partitioning $p_0$. 
}
   \label{Equivalence}
\newcommand{\LIF}{\lIf}
\begin{algorithmic}
\SetEndCharOfAlgoLine{}

\STATE \textbf{method} update\_records($q,h,q_\hypoaut,w$):
	\STATE Visitors$(q)\leftarrow$ Visitors$(q)\cup\{h\}$, 
	\STATE Paths$(h)\leftarrow w$
	\STATE Association$(q) \leftarrow (q_\hypoaut)$
	\STATE Push(New,$\{ h\} $)
\STATE \textbf{end method}	
 
\STATE 

\STATE \textbf{method} handle\_cluster\_conf($q,q_{\hypoaut},q'_{\hypoaut}$):
		\STATE  \textbf{find} $s\in\Sigma^*$ \textbf{s.t.} $\fhypoaut(q_\hypoaut,s)\neq \fhypoaut(q'_\hypoaut,s)$
		\FOR{$h\in$Visitors($q$) }
				\STATE $w\leftarrow Paths(h)\cdot s$
				\STATE \LIF {$f_R(w)\neq f_\hypoaut(w)$} {
				\RETURN{Reject, $w$}\Indp}
		\ENDFOR
		\STATE $p\leftarrow \mathit{ref}(p,h',$Visitors($q'$)$\setminus\{h'\}$)
		\RETURN{ Restart\_Exploration}
\STATE \textbf{end method}	

\STATE

\STATE \textbf{method} parallel\_explore($R,\hypoaut,p$):
\STATE empty all of: $Q,F,\delta$, New, Visitors, Paths,  Association
\STATE $q_0 \leftarrow p(h_0)$
\STATE update\_records($q_0,h_0,q_\hypoaut,_0,\varepsilon$)

\WHILE{New $\neq \emptyset$}
		\STATE $h \leftarrow$ Pop(New)
		\STATE $q \leftarrow p(h)$
		\STATE $q_\hypoaut \leftarrow$ Association$(q)$
		\STATE \LIF{$f_R(h)\neq \fhypoaut(q_\hypoaut)$} {
				\RETURN{Reject, (Paths($h$))} \Indp}
		\Indm
		\STATE \LIF{$q\in Q$} {\textbf{continue}}
			\STATE $Q \leftarrow Q\cup \{q\}$
			\STATE \LIF{$f_R(h)=Acc$} {$F \leftarrow F\cup \{q\}$}
			\FOR{$\sigma \in \Sigma$}
					\STATE $h'\leftarrow g_R(h,\sigma)$
					\STATE $q' \leftarrow p(h')$
					\STATE $\delta(q,\sigma)\leftarrow q'$
					\STATE \LIF{ $q'\in Q$ and \emph{Association}$(q')\neq \delta_\hypoaut(q_\hypoaut,\sigma)$} { \RETURN{ handle\_cluster\_conf($q,q_\hypoaut,\delta_\hypoaut(q_\hypoaut,\sigma)$)}\Indp}
					\Indm
					\STATE update\_records($q',h',\delta_\hypoaut(q_\hypoaut,\sigma)$,Paths$(h)\cdot\sigma$)
			\ENDFOR
\ENDWHILE
\RETURN{Accept}
\STATE \textbf{end method}

\STATE

\STATE \textbf{method} check\_equivalence($R,\hypoaut,p_0$):
\STATE $p\leftarrow p_0$
\STATE verdict $\leftarrow$ Restart\_Exploration
\WHILE{verdict = Restart\_Exploration}
\STATE verdict, $w \leftarrow$ parallel\_explore($R,\hypoaut,p$)
\ENDWHILE
\RETURN{verdict,$w$}
\STATE \textbf{end method}

\end{algorithmic}
\end{algorithm}

\section{Abstraction and Refinement}\label{Se:Quantisation}

Given a partitioning $p$, 
an R-state $h$, 
and a set of R-states $H\subseteq S\setminus\{h\}$, 
we 
refine $p$ in accordance with the requirements described in
\cref{Se:Resolution}.
We want to generalize the information given by $h$ and $H$ well, 
so as not to invoke excessive refinements. 
We also need an initial partitioning $p_0$ from which to start.

Our method is unaffected by 
the length of the R-states, and very conservative: 
each refinement increases the number of A-states by exactly one. 
Our experiments show that it is fast enough to quickly find counterexamples to proposed DFAs.

\subsection{Initial Partitioning}

As we 
wish to keep the 
abstraction as small as possible, 
we begin with no state separation at all: $p_0 : h \mapsto 0$. 

\subsection{Support-Vector based Refinement}

In this section we assume $p(h')=p(h)$ for every $h'\in H$, which is true for our case. 
The method generalizes trivially to cases 
where this is not true.\footnote{By removing 
from $H$ any vectors $h'$ for which $p(h')\neq p(h)$.}

We would like to allocate a region around the R-state $h$ that is
large enough to contain other R-states that behave similarly, 
but separate from neighboring R-states that 
do not. 
We achieve this by fitting an SVM~\cite{SVMOriginal} classifier with an RBF
kernel\footnote{While we see this as a natural choice, 
other kernels or classifiers may yield similar results. 
We do not explore such variations in this work.}
 to separate 
 $h$ from 
 $H$.
The max-margin property of the SVM ensures 
a large space around $h$, 
while the Gaussian RBF kernel allows for 
a non-linear partitioning of the space.

We use this classifier to split the A-state $p(h)$,
yielding a new partitioning $p'$ with exactly one more A-state than $p$. 
We track the refinements 
by arranging the obtained SVMs
in a decision tree,
where each node's decision is the corresponding SVM, 
and the leaves represent the current A-states.

Barring failure of the SVM, 
this approach satisfies the requirements of refinement operations, 
and avoids state explosion by adding only one A-state per refinement. 
Otherwise, the method fails to satisfy requirement 1. 
Nevertheless, at least one of the R-states $h'\in H$ is separated from $h$, and
later explorations can invoke further refinements if necessary. 
In practice this does not hinder the goal of the abstraction: finding counterexamples to equivalence queries.

The abstraction's storage is linear in the number of A-states it can map to; and computing an R-state's associated A-state may be linear in this number as well. 
However, as this number of A-states also grows very slowly (linearly in the number of refinements), this does not become a problem.

\subsection{Practical Considerations}
As the initial partitioning and 
the refinement operation are very coarse, 
the method may
accept very small but wrong DFAs.
To counter this, two measures are taken: 
\vspace{-0.2cm}
\begin{enumerate}
\setlength{\itemsep}{0pt}
\item One accepting and one rejecting sequence are provided to the teacher as potential counterexamples to be considered at every equivalence query. 
\item The first refinement uses an aggressive approach that generates a great (but manageable) number of A-states.
\end{enumerate}
\vspace{-0.2cm}
The first measure, necessary to prevent termination on a single state automaton,
requires only two samples.
These can be found by random sampling, or taken from the training set.\footnote{If 
no such samples exist, 
a single state DFA may be correct.}
In keeping with the observation made in \cref{Se:Resolution}, 
we take the shortest available samples.
The second measure prevents the extraction from too readily terminating on small DFAs. 
Our method for it is presented in \cref{maxgap}.

\subsubsection{Aggressive Difference-based Refinement}\label{maxgap}
We split $h$ from at least one of the vectors $h'\in H$ 
by splitting $S$ along the $d$ dimensions with the largest gap 
between $h$ and the mean $h_m$ of $H$,
down the middle of that gap. This 
refinement can be comfortably maintained in a decision tree, 
generating at the split point a tree of depth $d$ for which, 
on each layer $i=1,2,...,d$, 
each node is 
split along the dimension with the $i$-th largest gap.
This refinement follows intuitively from the quantization suggested by Omlin and Giles, 
but focuses only on the dimensions with the greatest deviation of values between the states being split 
and splits the `active' range of values.

The value $d$ may be set by the user, 
and increased if the extraction is suspected to have converged too soon. 
We found that values of around 7-10 
generally provide a strong enough initial partitioning of $S$, 
without making the abstraction too large for feasible exploration.

\section{Experimental Results}\label{Se:Results}

We demonstrate the effectiveness of our method 
on networks 
trained on 
the Tomita grammars (\citeyear{tomita82}),\footnote{The
Tomita grammars are the following 7 languages over the alphabet $\{0,1\}$: 
[1] \texttt{1$^*$},
[2] \texttt{(10)$^*$},
[3] the complement of \texttt{((0|1)$^*$0)$^*$1(11)$^*$(0(0|1)$^*$1)$^*$0(00)$^*$(1(0|1)$^*$)$^*$},
[4] all words $w$ not containing \texttt{000},
[5] all $w$ for which $\#_0(w)$ and $\#_1(w)$ are even (where $\#_a(w)$ is the number of $a$'s in $w$),
[6] all $w$ for which ($\#_0(w)-\#_1(w))\equiv_3 0$, and
[7] \texttt{0$^*$1$^*$0$^*$1$^*$}.
} 
used as benchmarks in previous automata-extraction work
\cite{NNExtractionGiles2017},
and on substantially more complicated languages.
We show the effectiveness of our 
equivalence query approach
over simple random sampling and present cases in which our method extracts informative DFAs whereas other approaches fail.
In addition, for some seemingly perfect networks, 
we find that our method quickly returns 
counterexamples representing deviations from the target language.

On all networks, 
we applied our method with
initial refinement depth $10$.
Unlike other extraction methods, 
where parameters must be tuned to find
the best DFA, 
no parameter tuning was required to achieve our results.

We clarify that when we refer to extraction time for any method, we consider the \emph{entire} process: from the moment the extraction begins, to the moment a DFA is returned.\footnote{Measured using \texttt{clock()}, of Python's \texttt{time} module, and covering among others: abstraction exploration, abstraction refinements (including training SVM classifiers), and \lstar refinements.}

\vspace{-0.5em}
\paragraph{Prototype Implementation and Settings}

We implemented all methods in Python, 
using PyTorch~\cite{pytorch} and  
scikit-learn~\cite{scikit-learn}.
For the SVM classifiers, 
we used the SVC variant, 
with regularization factor $C=10^4$ to encourage perfect splits and otherwise default parameters---in particular, 
the RBF kernel 
with gamma value $1/(\textrm{num features})$.

\vspace{-0.5em}
\paragraph{Training Setup}
As our focus was extraction, we trained all networks to $100\%$ accuracy on their train sets, and of these we considered only those that reached $99.9{+}\%$ accuracy on a dev set consisting of up to 1000 uniformly sampled words of each of the lengths $n\in{1,4,7,...,28}$. The positive to negative sample ratios in the dev sets were not controlled.

The train sets contained samples of various lengths, with a 1:1 ratio between the positive and negative samples from each length where possible. To achieve this, a large number of words were uniformly sampled for each length. When not enough samples of one class were found, we limited the ratio to 50:1, or took at most 50 samples if all were classified identically.
The train set sizes, and the lengths of the samples in them, are listed for every language in this paper in the supplementary material.

For languages where the positive class was unlikely to be found by random sampling---e.g. balanced parentheses or emails---we generated positive samples using tailored functions.\footnote{For instance, a function that creates emails by uniformly sampling 2 sequences of length $2{-}8$, choosing uniformly from the options \texttt{.com}, \texttt{.net}, and all \texttt{.co.XY} for \texttt{X,Y} lowercase characters, and then concatenating the three with an additional \texttt{@}.} In these cases we also generated negative samples by mutating the positive examples.\footnote{By adding, removing, changing, or moving up to 9 time characters.} 
Wherever a test set is mentioned, it was taken as a 1:1 sample set from the same distribution generating the positive and negative samples.

\vspace{-0.5em}
\paragraph{Effectiveness on Random Regular Languages}\label{Se:SmallRegular}

We first evaluated our method on the 7 Tomita grammars. 
We trained one 2-layer GRU network with hidden size $100$ 
for each grammar (7 RNNs in total).
All but one RNN reached $100\%$ dev accuracy; the one trained on the 6th Tomita grammar
reached $99.94\%$. 
For each RNN, 
our method correctly extracted and accepted the
target grammar in under 2 seconds.

The largest Tomita grammars have 5-state DFAs over a 2-letter alphabet. We also explored substantially more complex grammars: we trained 2-layer GRU networks with 
varying hidden-state sizes on 
10-state minimal DFAs generated randomly over a 3-letter alphabet. 
We applied our method to these networks
with a 30 second time limit---though most reached equivalence sooner. 
Extracted DFAs were compared against their networks on their
train sets and on $1000$ random samples for each of several
word-lengths. 

\cref{MainTable} shows the results. 
Each row represents 3 experiments:
$9$ random DFAs were generated, 
trained on, and extracted. The extracted DFAs are small, and highly accurate even on long sequences (length $1000$). 
Additional results showing similar trends,
including experiments on LSTM networks,
are available in the supplementary material.

\begin{table}[t]
\centering
\small
\caption{Accuracy  
of DFAs extracted from GRU networks 
representing small regular languages. 
Single values represent the average of 3 experiments, 
multiple values list the result for each experiment. 
Extraction time of 30 seconds is a timeout.}
	\label{MainTable}

\begin{scalebox}{0.85}{
\begin{tabular}{l||r||r||r|r|r|r|r}
      Hidden &  & DFA & \multicolumn{5}{c}{ Average Accuracy on Length }\\
Size      & Time (s)      &  Size   & 10 & 50 & 100 & 1000 & Train \\
\hline
50 & 30, 30, 30 & 11,11,155 & 99.9 & 99.8 & 99.9 & 99.9 & 99.9 \\
100 &11.0 & 11,10,11 & 100 & 99.9 & 99.9 & 99.9 & 100 \\
500 &30, 30, 30 & 10,10,10 & 100 & 99.9 & 100 & 99.9 & 100.0 \\

\vspace{-0.8cm}
\end{tabular}}
\end{scalebox}
\\

\end{table}

\vspace{-0.5em}
\paragraph{Comparison with a-priori Quantization}

In their 1996 paper, Omlin and Giles suggest 
partitioning the network state space by dividing each 
state dimension into $q$ equal intervals,
with $q$ being the \emph{quantization level}.
We tested this method on each of our networks, 
with $q=2$
and a time limit of $1000$ seconds
to avoid excessive memory consumption. 

In contrast to our method, which extracted on these same networks small and accurate DFAs within $30$ seconds, we found that for this method
this was not enough time to extract a complete DFA. 
The extracted DFAs were also very large---often 
with over $60{,}000$ states---and 
their coverage of sequences of length $1000$ tended to zero. 
For the covered sequences however, 
the extracted DFA's accuracy was 
often very high ($99$+$\%$), 
suggesting that quantization---while impractical---is 
sufficiently expressive to 
describe a network's state space. 
However, it is also possible that the sheer size of the quantization ($2^{50}$ for our smallest RNNs) simply allowed each explored R-state its own A-state, giving high accuracy by observation bias.

This highlights the 
key strength
of our method: 
in contrast to other methods, our method is able to find small and accurate DFAs representing a given RNN, when such DFAs are available. 
It does this in a fraction of the time required by other methods to complete their extractions.
This is because, unlike other methods,
it maintains from a very early point in extraction
a complete DFA that constitutes a continuously improving 
approximation of $R$.

\vspace{-0.5em}
\paragraph{Comparison with Random Sampling For Counterexample Generation}\label{Se:BP}
We show that there is merit to our approach to equivalence queries over simple random sampling.

Networks $R$ for which the ratio between 
accepting and rejecting sequences is very uneven may be 
closely approximated by simple DFAs---making 
it hard to differentiate between them and 
their \lstar proposed automata by random sampling. 
We trained two networks on one such language:
balanced parentheses (BP) over 
the 28-letter alphabet \texttt{\{a$,$b$,...,$z$,$($,$)\}} 
(the language of all sequences $w$ over \texttt{a-z()} 
in which
every opening parenthesis is eventually followed by 
a single corresponding closing parenthesis, 
and vice versa).
The networks were trained
to $100\%$ accuracy on train sets
of size ${\sim}44600$, 
containing samples with balanced parentheses up to depth $11$.
The two train sets had $36\%$ and $43\%$ negative samples, 
which were created by slightly mutating the positive samples.
The networks were a 2-layer GRU and a 2-layer LSTM, 
both with hidden size 50 per cell. 

We extracted from these networks using \lstarnogap, 
approaching equivalence queries
either with our method or by random sampling. 
We implemented the random sampling teacher to sample up to 1000 words of each length in increasing order.
For fairness, 
we also provided it with 
the same two initial samples our teacher was given, 
allowing it to check and possibly return them 
at every equivalence query.

We ran each extraction with a time limit of $400$ seconds
and found a nice pattern: 
every DFA proposed by \lstar 
represented BP to some bounded nesting depth,
and every counterexample taught it to increase that depth by 1.

\begin{table}
\centering
\caption{Accuracy and maximum nesting depth of extracted automata for networks trained on BP, using either abstractions (``Abstr") or random sampling (``RS'') for equivalence queries. Accuracy is measured with respect to the trained RNN.}\label{ExtractionScores}

\begin{tabular}{l||r|r||r|r}
& \multicolumn{2}{c||}{Train Set Accuracy} & \multicolumn{2}{|c}{Max Nest. Depth} \\
Network & Abstr & RS & Abstr & RS   \\

\hline
GRU & 99.98 & 87.12 & 8 & 2\\
LSTM & 99.98 & 94.19& 8 &3\\
\multicolumn{5}{c}{}
\end{tabular}

\caption{Counterexamples generated during extraction of automata from a GRU network trained on BP.}\label{BPExtraction}

\begin{tabular}{|l|r|l|r|}
\hline
\multicolumn{2}{|c|}{Refinement Based} & \multicolumn{2}{|c|}{Brute Force}\\
example & Time (s) & example & Time (s) \\

\hline
)) & 1.1 & )) & 0.4 \\
(()) & 1.2 & (()i)ma & 32.6\\
((())) & 2.1 &  &  \\
(((()))) & 3.1 & & \\
((((())))) & 3.8 & & \\
(((((()))))) & 4.4 & & \\
((((((())))))) & 6.6 & & \\
(((((((()))))))) & 9.2 & & \\
((((((((v()))))))) & 10.7 & & \\
((((((((a()z))))))))) & 8.3 & & \\
\hline
\end{tabular}

\end{table}

The accuracy of the extracted DFAs on the network train sets is shown in \cref{ExtractionScores}, 
along with the maximum depth the \lstar DFAs reached while still mimicking BP. 
For the GRU extractions,
the counterexamples and their generation times are listed in \cref{BPExtraction}. 
Note the speed and succinctness of those generated by our method 
as opposed to those generated by random sampling.

\vspace{-0.5em}
\paragraph{Adversarial Inputs} Excitingly, the penultimate counterexample returned by our method is 
an adversarial input: a sequence with unbalanced parentheses that the network (incorrectly) accepts.
This input is found in spite of the network's seemingly perfect behavior on its $44000$+ sample train set. 
We stress that the random sampler did not manage to find such samples.

Inspecting the extracted automata indeed reveals an almost-but-not-quite correct DFA for the BP language (the automata as well as the counterexamples are available in the supplementary material). The RNN overfit to random peculiarities in the training data and did not learn the intended language.

\vspace{-0.5em}
\paragraph{k-Means Clustering} We also implemented a simple k-means clustering and extraction approach and applied it to the BP networks with a variety of $k$ values, allowing it to divide the state space into up to $100$ clusters based on the states observed with the networks' train sets. This failed to learn any BP to any depth for either network: for both networks, it only managed to extract DFAs almost resembling BP to nesting depth 3 (accepting also some unbalanced sequences). 

\vspace{-0.5em}
\paragraph{Limitations}

Due to \lstarnogap's polynomial complexity and intolerance to noise,
for networks with complicated behavior, 
extraction becomes extremely slow and returns large DFAs.
Whenever applied to an RNN that has failed to generalize properly to its target language,
our method soon finds several adversarial inputs, builds a large DFA, and times out while refining it.\footnote{This happened also to our BP LSTM network, which timed out during \lstar refinement after the last counterexample.}

This does however demonstrate the ease with which the method 
identifies incorrectly trained 
networks.
These
cases are annoyingly frequent: 
for many RNN-acceptors with 100\% train and test accuracy on large test sets,
our method was able to find many simple misclassified examples.

For instance, 
for a seemingly perfect LSTM network trained on the regular expression 
$$\textrm{[a-z][a-z0-9]*@[a-z0-9]+.(com$|$net$|$co.[a-z][a-z])\$} $$
(simple email addresses over the 38 letter alphabet \texttt{$\{$a-z$,$0-9$,$@$,$.$\}$})  
to 100\% accuracy on a 40,000 sample train set and a 2,000 sample test set, 
our method quickly returned the counterexamples seen in \cref{fuzzycounterexamples},
showing clearly words that the network misclassified 
(e.g., 
\texttt{25.net}). 
We ran extraction on this network for 400 seconds, 
and while we could not extract a representative DFA in this time,\footnote{A 134-state DFA $\hypoaut$ was proposed by \lstar after 178 seconds, and the next refinement to $\hypoaut$ (4.43 seconds later) timed out. The accuracy of the 134-state DFA on the train set was nearly random. We suspect that the network learned such a complicated behavior that it simply could not be represented by any small DFA.}
our method did show that the network learned
a far more elaborate (and incorrect) function than needed.
In contrast, 
given a 400 second overall time limit, 
the random sampler did not find any counterexample
beyond the provided one.

We note that 
our implementation of kmeans clustering and extraction 
had no success with this network, 
returning a completely rejecting automaton 
(representing the empty language), 
despite trying $k$ values of up to $100$
and using all of the network states reached using
a train set with 50/50 ratio between 
positive and negative samples. 

Beyond demonstrating the 
capabilities of our method, 
these results also highlight the brittleness in
generalization of trained RNNs, 
and suggest that evidence based on test-set performance 
should be interpreted with extreme caution.

This reverberates the results of 
Gorman and Sproat (\citeyear{NNBad}),
who trained a neural architecture based on a multi-layer LSTM to mimic a finite state transducer (FST) 
for number normalization. 
They showed that the RNN-based network, 
trained on 22M samples and validated on a 2.2M sample development set 
to 0\% error on both, 
still had occasional errors (though with error rate $<$ 0.0001) 
when applied to a 240,000 sample blind test set.

\begin{table}

\caption{Counterexamples generated during extraction from an LSTM email network with $100\%$ train and test accuracy. 
Examples of the network deviating from its target language 
are shown in bold.} \label{fuzzycounterexamples}
\begin{scalebox}{0.9}{
\begin{tabular}{|l|r|r|r|}
\hline
Counter- & & Network  & Target\\
example & Time (s) & Classification & Classification\\

\hline
0@m.com & provided & $\surd$ & $\surd$ \\
@@y.net & 2.93 & $\times$ & $\times$ \\
\textbf{25.net} & 1.60 & $\surd$ & $\times$ \\
\textbf{5x.nem} & 2.34 & $\surd$ & $\times$ \\
0ch.nom & 8.01 &$\times$ &$\times$ \\
9s.not & 3.29 & $\times$& $\times$ \\
\textbf{2hs.net} & 3.56 & $\surd$&$\times$ \\
@cp.net & 4.43 &$\times$ &$\times$ \\
\hline
\end{tabular}}
\end{scalebox}

\vspace{-0.3cm}
\end{table}

\section{Conclusions}\label{Se:Conc}
We present a novel technique for extracting deterministic finite automata from recurrent neural networks 
with roots in exact learning. 
As our method makes no assumptions as to the internal configuration of the network, 
it is easily applicable to any RNN architecture,
including the popular LSTM and GRU models.

In contrast to previous methods, our method is not affected by hidden state-size, and 
successfully extracts representative DFAs for 
any networks that
can indeed be represented as such. 
Unlike other extraction approaches, our technique works with little to no parameter tuning, and requires very little prior information to get started 
(the input alphabet, and 2 labeled examples).

Our method is guaranteed to never extract a DFA 
more complicated than the language of the RNN being considered. 
Moreover, the counterexamples returned during our extraction can 
point us to incorrect patterns 
the network has learned without our awareness.

Beyond scalability and ease of use, 
our method can return reasonably accurate DFAs even if extraction is cut short.
Moreover,
we have shown that for networks that do correspond to succinct automata, 
our method gets very good results---generally 
extracting small, succinct DFAs 
with accuracies of over $99\%$ with respect to 
their networks, 
in seconds or tens of seconds. 
This is in contrast to existing methods, 
which require orders of magnitude more time to complete, 
and often return large and cumbersome DFAs 
(with tens of thousands of states).

\clearpage
\appendix

\section*{Supplementary Material}
This supplementary material contains
 a description of the \lstar algorithm (\cref{App:Lstar}), 
 and additional experimental results and details (\cref{App:Results}).
\section{Angluin's L* Algorithm }\label{App:Lstar}

\begin{algorithm}
\caption{L* Algorithm with explicit membership and equivalence queries.}\label{lstarAlgo} 
\begin{algorithmic}

\STATE $S \leftarrow \{ \epsilon \}, E \leftarrow \{ \epsilon \}$
\FOR{$(s \in  S)$, $(a \in \Sigma)$, and $(e \in E)$}
\STATE $T[s, e] \leftarrow \textbf{Member}(s \cdot e)$
\STATE $T[s \cdot a, e] \leftarrow \textbf{Member}(s \cdot a \cdot e)$
\ENDFOR
\WHILE{True}
\WHILE{$(s_{new} \leftarrow Closed(S, E, T) \neq \bot$)}
\STATE $Add(S, s_{new} )$
\FOR{$(a \in \Sigma, e \in E)$} 
\STATE  $T[s_{new} \cdot a, e] \leftarrow \textbf{Member} (s_{new} \cdot a \cdot e)$
\ENDFOR
\ENDWHILE
\STATE $\hypoaut \leftarrow MakeHypothesis (S, E, T)$
\STATE $cex \leftarrow \textbf{Equivalence}(\hypoaut)$
 \IF{$cex = \bot$}
 \RETURN{$\hypoaut$}
 \ELSE
\STATE  $e_{new} \leftarrow FindSuffix (cex )$
\STATE  $Add(E, e_{new})$
\FOR{$(s \in S, a \in \Sigma)$} 
\STATE     $T[s, e_{new} ] \leftarrow \textbf{Member} (s \cdot e_{new} )$
\STATE    $T[s \cdot a, e_{new} ] \leftarrow \textbf{Member} (s \cdot a \cdot e_{new} )$
\ENDFOR
\ENDIF
\ENDWHILE
\end{algorithmic}
\end{algorithm}

Angluin's \lstar algorithm (\citeyear{Lstar}) is an exact learning algorithm for regular languages. The algorithm learns an unknown regular language $U$ over an alphabet $\Sigma$, generating a DFA that accepts $U$ as output.  We only provide a brief and informal description of the algorithm; for further details see~\cite{Lstar,AngluinInsights}.

\cref{lstarAlgo} shows the \lstar algorithm. This version is adapted from Alur et al. (\citeyear{JIST}), where the membership and equivalence queries have been made more explicit than they appear in Angluin (\citeyear{Lstar}).

The algorithm maintains an \emph{observation table} $(S,E,T)$ that records whether strings belong to $U$. In \cref{lstarAlgo}, this table is represented by the two-dimensional array $T$, with dimensions $|S| \times |E|$, where, informally, we can view $S$ as a set of words that lead from the initial state to states of the hypothesized automaton, and $E$ as a set of words serving as experiments to separate states. The table $T$ itself maps a word $w \in (S \cup S \cdot \Sigma) \cdot E$ to \texttt{True} if $w \in U$ and \texttt{False} otherwise.

The table is updated by invoking membership queries to the teacher. When the algorithm reaches a consistent and closed observation table (meaning that all states have outgoing transitions for all letters, without contradictions), the algorithm constructs a hypothesized automaton $\hypoaut$, and invokes an \emph{equivalence query} to check whether $\hypoaut$ is equivalent to the automaton known to the teacher. If the hypothesized automaton accepts exactly $U$, then the algorithm terminates. If it is not equivalent, then the teacher produces a counterexample showing a difference between $U$ and the language accepted by $\hypoaut$.

A simplified run through of the algorithm is as follows: 
the learner starts with an automaton with one state---the initial state---which is accepting or rejecting according to the classification of the empty word. 
Then, for every state in the automaton, for every letter in the alphabet, 
the learner verifies by way of membership queries that for every shortest sequence reaching that state, 
the continuation from that prefix with that letter is correctly classified. 
As long as an inconsistency exists, 
the automaton is refined. 
Every time the automaton reaches a consistent state 
(a complete transition function, with no inconsistencies by single-letter extensions), 
that automaton is presented to the teacher as an equivalence query. 
If it is accepted, the algorithm completes; otherwise, 
it uses the teacher-provided counterexample to further refine the automaton.

\section{Additional Results}\label{App:Results}
\subsection{Random Regular Languages}
\begin{table*}
\centering
\small
\caption{Results for DFA extracted using our method from 2-layer GRU and LSTM
	networks with various state sizes, trained on random regular languages of
	varying sizes and alphabets. Each row in each table represents 3 experiments
	with the same parameters (network hidden-state size, alphabet size, and minimal target DFA size). Single values represent the average of the 3 experiments, multiple values list the result for each experiment. An extraction time of 30 seconds signals a timed out extraction (for which the last automaton proposed by \lstar is taken as the extracted automaton).}
\label{MainTableFull}

\begin{scalebox}{0.81}{
\begin{tabular}{l||l||l||r||r||r|r|r|r|r}
\multicolumn{10}{c}{ }\\
\multicolumn{10}{c}{\textbf{Extraction from LSTM Networks --- Our Method}} \\
Hidden      & Alphabet       & Language /  & Extraction& Extracted  & \multicolumn{5}{c}{ Average Extracted DFA Accuracy }\\
Size       & Size & Target DFA Size & Time (s)      &  DFA Size   & $l$=10 & $l$=50 & $l$=100 & $l$=1000 & Training \\
\hline
50 & 3 & 5 & 2.9 & 5,5,6 & 100.0 & 99.96 & 99.86 & 99.90 & 100.0 \\
100 & 3 & 5 & 2.9 & 5,5,2 & 92.96 & 92.96 & 93.73 & 93.46 & 91.06 \\
500 & 3 & 5 & 11.8 & 5,5,5 & 100.0 & 100.0 & 100.0 & 99.96 & 100.0 \\
50 & 5 & 5 & 30, 30, 30 & 68, 59, 115 & 99.96 & 99.93 & 99.76 & 99.93 & 99.99 \\
100 & 5 & 5 & 30, 7.7, 30 & 57, 5, 38 & 99.96 & 99.96 & 99.96 & 99.90 & 100.0 \\
500 & 5 & 5 & 30, 20.7, 19.0 & 5, 5, 5 & 100.0 & 100.0 & 99.93 & 99.90 & 100.0 \\
50 & 3 & 10 & 30, 30, 11.1 & 10, 10, 10 & 99.96 & 99.96 & 99.90 & 99.90 & 100.0 \\
100 & 3 & 10 & 7.6, 30, 7.7 & 10, 10, 11 & 99.96 & 99.93 & 99.96 & 99.96 & 100.0 \\
500 & 3 & 10 & 30, 30, 30 & 10, 9, 10 & 92.30 & 92.80 & 93.70 & 93.43 & 92.30 \\
\end{tabular}}
\end{scalebox}

\begin{scalebox}{0.81}{
\begin{tabular}{l||l||l||r||r||r|r|r|r|r}
\multicolumn{10}{c}{ }\\
\multicolumn{10}{c}{\textbf{Extraction from GRU Networks --- Our Method}} \\
Hidden      & Alphabet       & Language / & Extraction& Extracted  & \multicolumn{5}{c}{ Average Extracted DFA Accuracy }\\
Size       & Size & Target DFA Size & Time (s)      &  DFA Size   & $l$=10 & $l$=50 & $l$=100 & $l$=1000 & Training \\
\hline
50 & 3 & 5 & 1.7 & 5,5,6 & 100.0 & 100.0 & 99.86 & 99.96 & 100.0 \\
100 & 3 & 5 & 4.1 & 5,5,5 & 100.0 & 100.0 & 100.0 & 99.96 & 100.0 \\
500 & 3 & 5 & 7.0 & 5,5,5 & 100.0 & 100.0 & 100.0 & 100.0 & 100.0 \\
50 & 5 & 5 & 30, 30, 8.2 & 150,93,5 & 100.0 & 99.90 & 99.93 & 99.86 & 100.0 \\
100 & 5 & 5 & 9.0, 8.0, 30& 5,5,16 & 100.0 & 100.0 & 99.96 & 99.96 & 99.99 \\
500 & 5 & 5 & 15.5, 30, 25.6 & 5,5,5 & 100.0 & 100.0 & 99.96 & 100.0 & 100.0 \\
50 & 3 & 10 & 30, 30, 30 & 11,11,155 & 99.96 & 99.83 & 99.93 & 99.93 & 99.99 \\
100 & 3 & 10 & 11.0 & 11,10,11 & 100.0 & 99.93 & 99.96 & 99.93 & 100.0 \\
500 & 3 & 10 & 30, 30, 30 & 10,10,10 & 100.0 & 99.93 & 100.0 & 99.90 & 100.0 \\
\multicolumn{10}{c}{ }\\
\end{tabular}}
\end{scalebox}
\\

\caption{Results for automata extracted using Omlin \& Giles' a-priori
	quantization as described in their \citeyear{NNExtraction} paper, with quantization level 2,
	from the same networks used in Table \ref{MainTableFull} (3 networks for each set of
	parameters and network type).}\label{BlindTableFull}

\begin{scalebox}{0.81}{
\begin{tabular}{l||l||l||rrr||rr|rr|rr|rr|rr|rr}
\multicolumn{18}{c}{ }\\
\multicolumn{18}{c}{\textbf{Extraction from LSTM Networks --- O\&G Quantization}} \\
Hidden  & Alphabet &Language/ & \multicolumn{3}{|c||}{Extracted}& \multicolumn{12}{c}{Coverage / Accuracy (\%)}\\
Size       & Size &  Target DFA Size & \multicolumn{3}{|c||}{DFA Sizes}&\multicolumn{2}{|c|}{$l$=1} &\multicolumn{2}{|c|}{$l$=5}&\multicolumn{2}{|c|}{$l$=10}&\multicolumn{2}{|c|}{$l$=15}&\multicolumn{2}{|c|}{$l$=50}&\multicolumn{2}{|c}{Training}\\
\hline
50 & 3 & 5 & 3109 & 3107&3107&100&100&100&100&30.66&83.65&3.87&81.53&0.0&NA&27.44&88.0\\
100 & 3 & 5 & 2225&2252&2275&100&100&100&100&7.57&80.50&0.07&50.0&0.0&NA&19.31&84.57\\
500 & 3 & 5 &585&601&584&100&100&100&100&0.0&NA&0.0&NA&0.0&NA&8.80&71.71\\
50 & 5 & 5 & 1956&1973&1962&100&100&100&73.93&0.03&100&0.0&NA&0.0&NA&12.39&78.34\\
100 & 5 & 5 & 1392&1400&1400&100&100&100&64.3&0.0&NA&0.0&NA&0.0&NA&11.19&74.80\\
500 & 5 & 5 & 359&366&366&100&100&33.43&70.60&0.0&NA&0.0&NA&0.0&NA&6.24&73.92\\
50 & 3 & 10 & 3135&3238&3228&100&100&100&100&29.43&83.72&4.57&94.19&0.0&NA&27.70&88.80\\
100 & 3 & 10 & 2294&2282&2272&100&100&100&100&0.90&91.30&0.0&NA&0.0&NA&16.83&81.07\\
500 & 3 & 10 & 586&589&589&100&100&100&100&0.0&NA&0.0&NA&0.0&NA&8.48&74.77\\
\end{tabular}}
\end{scalebox}

\begin{scalebox}{0.79}{
\begin{tabular}{l||l||l||rrr||rr|rr|rr|rr|rr|rr}
\multicolumn{18}{c}{ }\\
\multicolumn{18}{c}{\textbf{Extraction from GRU Networks --- O\&G Quantization}} \\
Hidden  & Alphabet & Language/  & \multicolumn{3}{|c||}{Extracted}& \multicolumn{12}{c}{Coverage / Accuracy (\%)}\\
Size       & Size & Target DFA Size & \multicolumn{3}{|c||}{DFA
Sizes}&\multicolumn{2}{|c|}{$l$=1} &\multicolumn{2}{|c|}{$l$=5}&\multicolumn{2}{|c|}{$l$=10}&\multicolumn{2}{|c|}{$l$=15}&\multicolumn{2}{|c|}{$l$=50}&\multicolumn{2}{|c}{Training}\\
\hline
50 & 3 & 5 & 4497& 4558& 4485 & 100&100 & 100&100 & 50.73&90.52 & 25.26&91.95 & 2.66&96.25 & 42.28&91.08 \\
100 & 3 & 5 & 3188& 3184&3197 & 100&100 & 100&100 & 3.50&66.66 & 0.07&50.0 & 0.0&NA & 19.14&83.63 \\
500 & 3 & 5 & 1200& 1221&1225 & 100&100 & 100&100 & 0.0&NA & 0.0&NA & 0.0&NA & 8.98&74.45 \\
50 & 5 & 5 & 2810& 2796& 2802 & 100&100 & 100&87.97 & 0.10&100 & 0.0&NA & 0.0&NA & 14.56&80.61 \\
100 & 5 & 5 & 1935& 1941& 1936 & 100&100 &100&73.17 & 0.0&NA & 0.0&NA & 0.0&NA & 12.05&76.39 \\ 
500 & 5 & 5 & 721&  706&749 & 100&100 & 91.03&55.94 & 0.0&NA & 0.0&NA & 0.0&NA & 9.52&71.53 \\
50 & 3 & 10 & 4598& 4582& 4586 & 100&100 & 100&100 & 15.73&79.76 & 1.23&70.71 & 0.0&NA & 24.32&86.93 \\
100 & 3 & 10 & 3203&3192& 3194 & 100&100 & 100&100 & 0.3&81.25 & 0.0&NA & 0.0&NA & 19.18&83.84 \\ 
500 & 3 & 10 & 1226&1209& 1209 & 100&100 & 100&100 & 0.0&NA & 0.0&NA & 0.0&NA & 13.39&76.64 \\
\multicolumn{18}{c}{ }\\
\end{tabular}}
\end{scalebox}

\end{table*}

We show results for extraction from GRU and LSTM networks with varying hidden sizes, trained on small regular languages of varying alphabet size and DFA size.
Each extraction was limited to 30 seconds, 
and had initial refinement depth 10.
For each of the combinations of state size and target language complexity, 
3 networks of each type were trained, 
each on a different (randomly generated) language. 
The full results of these experiments are shown in \cref{MainTableFull}. 
Note that each row in each of the tables represents 3 experiments, 
i.e. in total $9\times 2\times 3=54$ random DFAs were generated, 
trained on, and re-extracted. 

We note with satisfaction that in 36 of the 54 experiments, the extraction process reached equivalence on a regular language identical to the target language the network had been trained on. We also note that on one occasion in the LSTM experiments the extraction reached equivalence too easily, accepting an automaton of size 2 that ultimately was not a great match for the network. Such a problem could be countered by increasing the initial split depth, for instance when sampling shows that a too-simple automaton has been extracted.

We also ran extraction with Omlin and Giles' a-priori quantization on each of these networks, with quantization level 2 and a time limit of 50 seconds. The extractions did not complete in this time.
For completeness, we present the size, coverage, and accuracies of these partially extracted DFAs in \cref{BlindTableFull}. 
The smaller size of the extracted DFAs for networks with larger hidden size is a result of the transition function computation taking longer: this slows the BFS exploration and thus the extraction visits less states in the allotted time.

\subsection{Balanced Parentheses}
\begin{table*}
\centering
\caption{Extraction of automata from an LSTM network trained to 100\% accuracy on the training set for the language of balanced parentheses over the 28-letter alphabet \texttt{a-z$,$($,$)}. The table shows the counterexamples and the counterexample generation times for each of the successive equivalence queries posed by \lstar during extraction, for both our method and a brute force approach. Each successive equivalence query from \lstar was an automaton classifying the language of all words with balanced parentheses up to nesting depth $n$, with increasing $n$.}\label{BPExtractionLSTM}
\begin{tabular}{|l|r|l|r|}
\multicolumn{4}{c}{ }\\
\hline
\multicolumn{2}{|c|}{Refinement Based} & \multicolumn{2}{|c|}{Brute Force}\\
Counterexample & Time (seconds) & Counterexample & Time (seconds) \\
\hline
)) & 1.4 & )) & 1.5\\
(()) & 1.6 & tg(gu()uh) & 57.5\\
((())) & 3.1 & ((wviw(iac)r)mrsnqqb)iew & 231.5 \\
(((()))) & 3.1 &  &  \\
((((())))) & 3.4 & & \\
(((((()))))) & 4.7 & & \\
((((((())))))) & 6.3 & & \\
(((((((()))))))) & 9.2 & & \\
((((((((())))))))) & 14.0 & & \\
\hline
\end{tabular}

\end{table*}

\begin{figure}

\centering
\includegraphics[width=.15\textwidth]{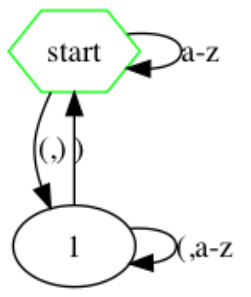}\hfill
\includegraphics[width=.25\textwidth]{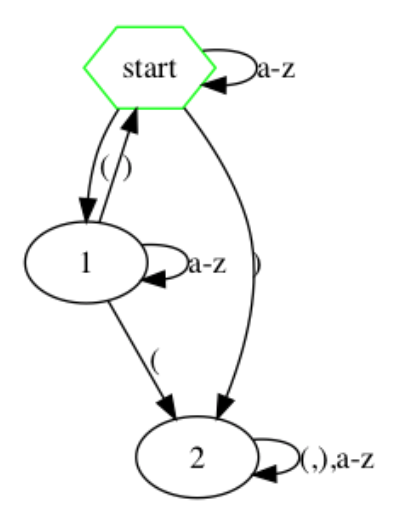}\hfill
\includegraphics[width=.25\textwidth]{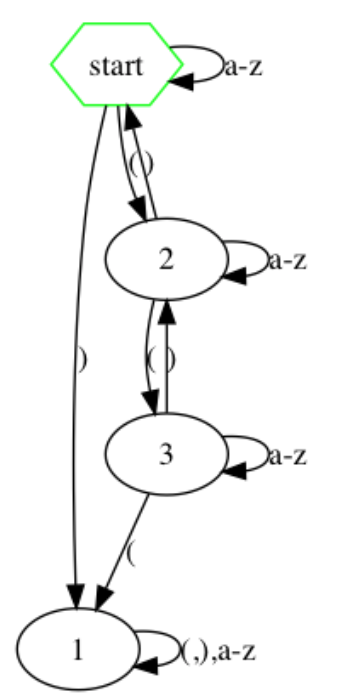}\hfill
\includegraphics[width=.2\textwidth]{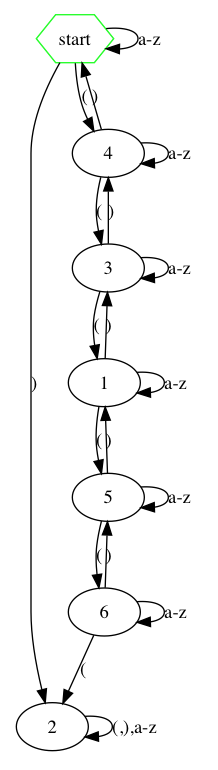}

\caption{Select automata of increasing size for recognizing balanced parentheses over the 28 letter alphabet \texttt{a-z$,$($,$)}, up to nesting depths 1 (flawed), 1 (correct), 2, and 5, respectively.\label{BPdepths}}

\end{figure}

\begin{figure*}

\centering
\includegraphics[width=.45\textwidth]{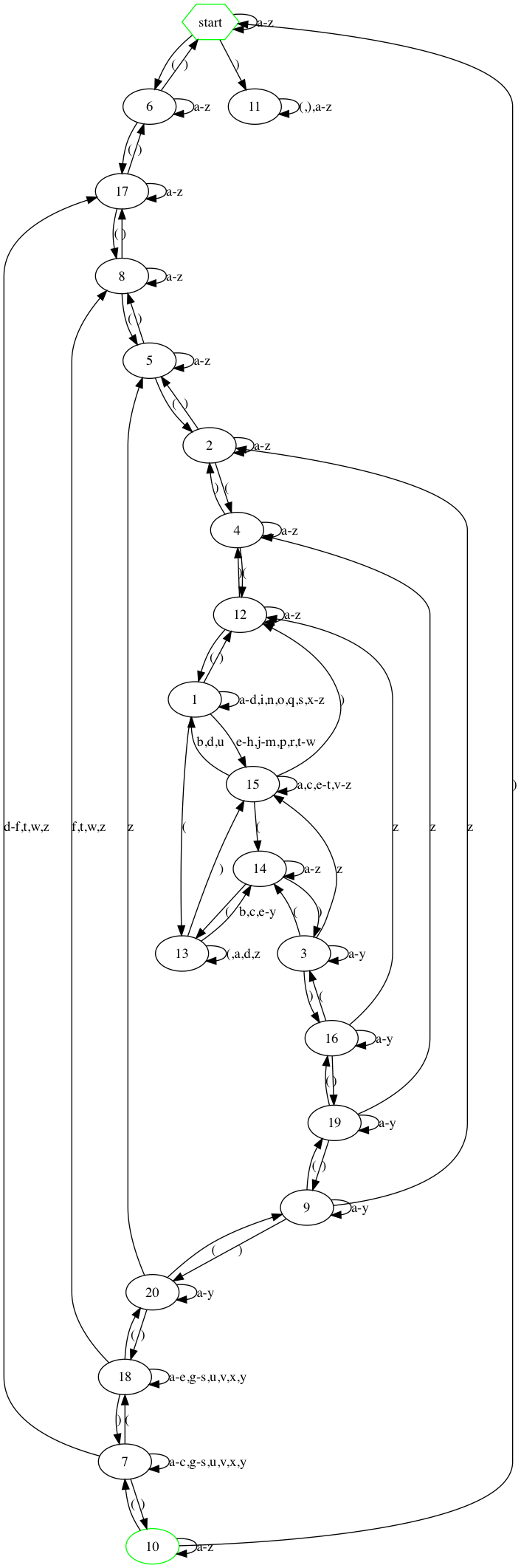}\hfill

\caption{Automaton no longer representing a language of balanced parentheses up to a certain depth. (Showing how a trained network may be overfitted past a certain sample complexity.) \label{BPbroken}}

\end{figure*}

We trained a GRU and an LSTM network on the irregular language of balanced parentheses, 
and then attempted to extract automata from these networks. 
For both, \lstar at first proposed a series of automata each representing the language of balanced parentheses to increasing nesting depth. 
Some of these are shown in \cref{BPdepths}. 
For the GRU network, after a certain point in the extraction, 
we even found a counterexample which showed the network had not generalized correctly to the language of balanced parentheses, 
and the next automaton returned resembled---but 
was not quite---an 
automaton for balanced parentheses nested to some depth. We show this automaton in \cref{BPbroken}.

In our main submission, we show the counterexamples returned during a 400-second extraction from the GRU network, as generated either by random sampling or by our method. For completeness, we present now in \cref{BPExtractionLSTM} the counterexamples for the LSTM extraction.

\subsection{Other Interesting Examples}
\subsubsection{Counting}
\begin{figure}

\centering
\includegraphics[width=.5\textwidth]{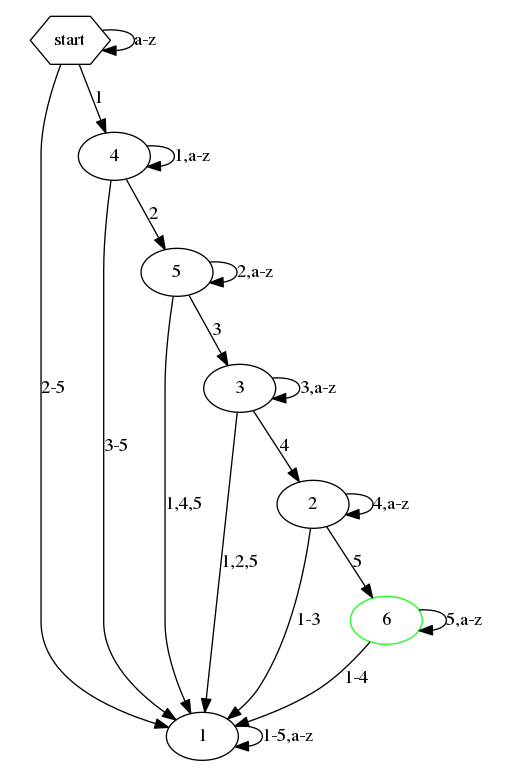}\hfill

\caption{DFA representing the regular language $ \textrm{[a-z]*1[a-z1]*2[a-z2]*3[a-z3]*4[a-z4]*5[a-z5]*\$}$ over the alphabet \texttt{\{a$,$b$,...,$z$,$1$,$2$,...,$5\}}}.\label{Im:Counting}

\end{figure}

\begin{table*}
		\caption{Counterexamples returned to the equivalence queries made by \lstar during extraction of a DFA from a network trained to 100\% accuracy on both train and test sets on the regular language $ \textrm{[a-z]*1[a-z1]*2[a-z2]*3[a-z3]*4[a-z4]*5[a-z5]*\$}$ over the 31-letter alphabet \texttt{\{a$,$b$,...,$d$,$1$,$2$,...,$5\}}. Counterexamples highlighting the discrepancies between the network behavior and the target behavior are shown in bold.}\label{CountingCounterexamples}
	\centering
	
	\begin{tabular}{|l|r|r|r|}
        \multicolumn{4}{c}{\bf Counterexample Generation for the Counting Language} \\
		\hline
		Counterexample & Generation Time (seconds) & Network Classification & Target Classification \\
		\hline
		12345 & provided & True & True\\
		512345 & 8.18 & False & False\\
		\textbf{aca11} & 85.41 & True & False\\
		\textbf{blw11} & 0.50 & True & False\\
		dnm11 & 0.96 & False & False\\
		bzm11 & 0.90 & False & False\\
		\textbf{drxr11} & 0.911 & True & False\\
		brdb11 & 0.90 & False & False\\
		\textbf{elrs11} & 1.16 & True & False\\
		hu11 & 1.93 & False & False\\
		ku11 & 2.59 & False & False\\
		ebj11 & 2.77 & False & False\\
		\textbf{pgl11} & 3.77 & True & False\\
		reeg11 & 4.16 & False & False\\
		eipn11 & 5.66 & False & False\\
		\hline
	\end{tabular}

\end{table*}

We trained an LSTM network with 2 layers and hidden size 100 (giving overall state size $d_s=2\times2\times100=400$) on the regular language 
$$ \textrm{[a-z]*1[a-z1]*2[a-z2]*3[a-z3]*4[a-z4]*5[a-z5]*\$}$$ 
over the 31-letter alphabet \texttt{\{a$,$b$,...,$z$,$1$,$2$,...,$5\}}, i.e. the regular language of all sequences \texttt{1+2+3+4+5+} with lowercase letters \texttt{a-z} scattered inside them. We trained this network on a train set of size 20000 and tested it on a test set of size 2000 (both evenly split on positive and negative examples), and saw that it reached 100\% accuracy on both.

We extracted from this network using our method. Within 2 counterexamples (the provided counterexample \texttt{12345}, and another generated by our method), and after a total of 9.5 seconds, \lstar proposed the automaton representative of the network's target language, shown in \cref{Im:Counting}. However, our method did not accept this DFA as the correct DFA for the network. Instead, after a further 85.4 seconds of exploration and refinement, the counterexample \texttt{aca11} was found and returned to \lstarnogap, meaning: our method found that the network accepted the word \texttt{aca11} --- despite this word not being in the target language of the network and the network having 100\% accuracy on both its train and test set.

Ultimately, after 400 seconds our method extracted from the network (but did not reach equivalence on) a DFA with 118 states, returning the counterexamples listed in \cref{CountingCounterexamples} and achieving 100\% accuracy against the network on its train set, and 99.9+\% accuracy on all sampled sequence lengths. 
We note that by the nature of our method, 
the complexity of this DFA is necessarily
an indicator of the inherent complexity of 
the concept to which the trained network has generalized.
\subsubsection{Tokenized JSON Lists}

\begin{figure*}

\centering
\begin{subfigure}{0.4\textwidth}
	\includegraphics[width=\textwidth]{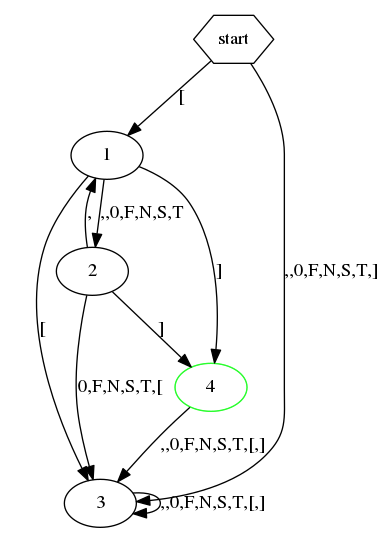}\hfill
	\caption{}\label{json1}
\end{subfigure}
\begin{subfigure}{0.4\textwidth}
	\includegraphics[width=\textwidth]{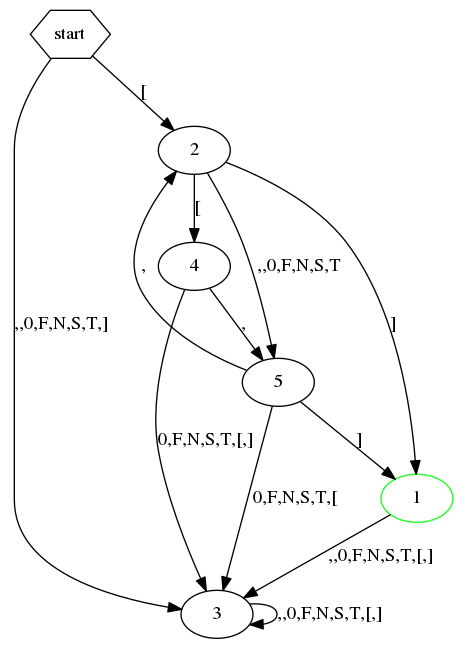}\hfill
	\caption{}\label{json2}
\end{subfigure}
\caption{Two DFAs resembling, but not perfectly, the correct DFA for the regular language of tokenized JSON lists, $\textrm{(\textbackslash[\textbackslash])$|$(\textbackslash[[S0NTF](,[S0NTF])*\textbackslash])\$}$. DFA \ref{json1} is almost correct, but accepts also list-like sequences in which the last item is missing, i.e., there is a comma followed by a closing bracket. DFA \ref{json2} is returned by \lstar after the teacher (network) rejects \ref{json1}, but is also not a correct representation of the target language --- treating the sequence \texttt{[,} as a legitimate list item equivalent to the characters \texttt{S$,$0$,$N$,$T$,$F} }.

\end{figure*}

\begin{table*}
\centering
\caption{Counterexamples returned to the equivalence queries made by \lstar during extraction of a DFA from a network trained to 100\% accuracy on both train and test sets on the regular language $\textrm{(\textbackslash[\textbackslash])$|$(\textbackslash[[S0NTF](,[S0NTF])*\textbackslash])\$} $ over the 8-letter alphabet \texttt{\{[$,$]$,$S$,$0$,$N$,$T$,$F$,$,\}}. Counterexamples highlighting the discrepancies between the network behaviour and the target behaviour are shown in bold.}\label{JsonCounterexamples}

\begin{tabular}{|l|r|r|r|}
        \multicolumn{4}{c}{\bf Counterexample Generation for the Non-Nested Tokenized
        JSON-lists Language} \\
\hline
Counterexample & Generation Time (seconds) & Network Classification & Target Classification \\
\hline
[]&provided&True&True\\
{[SS]}&3.49&False&False\\
\textbf{[[,]}		&	7.12	&		True	&	False			\\
\textbf{{[S,,}}		&	8.61	&	True	&			False			\\
\textbf{[0,F}		&	8.38	&True	&			False			\\
{[N,0,}					&	8.07	&	False	&		False			\\
\textbf{{[S,N,0,}}	&	9.43	&	True	&			False			\\
{[T,S,}					&	9.56	&False		&		False			\\
{[S,S,T,[]}				&	15.15	&	False	&			False		\\	
{[F,T,[}					&	3.23	&	False			&	False			\\
\textbf{{[N,F,S,0}}	&	10.04	&	True		&		False		\\	
\textbf{{[S,N,[,,,,}}	&	27.79	&	True			&	False		\\	
\textbf{{[T,0,T,	}}	&	28.06	&True	&			False			\\
\textbf{{[S,T,0,],}}	&	26.63	&	True&				False		\\	
\hline
\end{tabular}
\end{table*}

We trained a GRU network with 2 layers and hidden size 100 on the regular language representing a simple tokenized JSON list with no nesting, $$ \textrm{(\textbackslash[\textbackslash])$|$(\textbackslash[[S0NTF](,[S0NTF])*\textbackslash])\$} $$ over the 8-letter alphabet \texttt{\{[$,$]$,$S$,$0$,$N$,$T$,$F$,$,\}}, to accuracy 100\% on a training set of size 20000 and a test set of size 2000, both evenly split between positive and negative examples. As before, we extracted from this network using our method.

Within 2 counterexamples (1 provided and 1 generated) and a total of 3.8 seconds, our method extracted the automaton shown in \cref{json1}, which is almost but not quite representative of the target language. 7.12 seconds later it returned a counterexample to this DFA which pushed \lstar to refine further and return the DFA shown in \cref{json2}, which is also almost but not quite representative of zero-nesting tokenized JSON lists. 

Ultimately, after 400 seconds, our method extracted (but did not reach equivalence on) an automaton of size 441, returning the counterexamples listed in \cref{JsonCounterexamples} and achieving 100\% accuracy against the network on both its train set and all sampled sequence lengths. As before, we note that each state split by the method is justified by concrete inputs to the network, and so the extraction of a large DFA is a sign of the inherent complexity of the learned network behavior.

\subsection{Train Set Details}

\begin{table*}
		\caption{Train set statistics for networks used in this work. The random regular language networks used in the main work were based on minimal DFAs of size 10 over alphabets of size 3, with 3 languages per hidden state size. We list statistics for their train sets in grouped by the hidden size. The train set sizes and lengths were the same for each of these random languages, but the number of positive/negative samples found each time varied slightly.}
		\label{TrainSetSizes}
	\centering
	
	\begin{tabular}{|l|c|c|r|r|r|}
        \multicolumn{6}{c}{\bf Train Set Stats} \\
		\hline
		Language & Architecture & Hidden Size & Train Set Size & Of Which Positive Samples & Lengths in Train Set \\
		\hline
		Tomita 1 & GRU & 100 & 613 & 14 & 0-13,16,19,22\\
		Tomita 2 & GRU & 100 &  613 & 8 & 0-13,16,19,22\\
		Tomita 3 & GRU & 100 & 2911 & 1418 & 0-13,16,19,22\\
		Tomita 4 & GRU & 100 & 2911 & 1525 & 0-13,16,19,22\\
		Tomita 5 & GRU & 100 & 1833 & 771 & 0-13,16,19,22\\
		Tomita 6 & GRU & 100 & 3511 & 1671 & 0-13,15-20\\
		Tomita 7 & GRU & 100 & 2583 & 1176 & 0-13,16,19,22\\
		Random 1-3 & GRU & 50 & 16092 & 8038, 7768, 8050 & 1-15,16,18,...,26\\
		Random 4-6 & GRU & 100 & 16092 & 7783, 7842, 8167 & 1-15,16,18,...,26\\
		Random 7-9 & GRU & 500 & 16092 & 8080, 8143, 7943 & 1-15,16,18,...,26\\
		Balanced Parentheses & GRU & 50 & 44697 & 25243 & 0-73\\
		Balanced Parentheses & LSTM & 50 & 44816 & 28781 & 0-81\\
		emails & LSTM & 100 & 40000 & 20000 & 0-34\\
		JSON Lists & GRU & 100 & 20000 & 10000 & 0-74\\
		Counting & LSTM & 100 & 20000 & 10000 & 0-43\\
		\hline
	\end{tabular}

\end{table*}

The sizes, sample lengths, and positive to negative ratios of the samples in the train sets are listed here (\cref{TrainSetSizes}) for each of the networks used in our main experiments, as well as for the JSON and Counting languages. 
Note that for some languages (such as the first Tomita grammar, \texttt{1$^*$}), there are very few positive/negative samples. For these languages, the test sets are less balanced between positive and negative samples.

We reiterate that all networks used in this work were trained to $100\%$ train set accuracy and reached at least $99.9\%$ on a set of 1000 samples from each of the lengths $4,7,10,...,28$.

An explicit description of the Tomita grammars can be found in \cite{tomita82}.

\section*{Acknowledgments}
The research leading to the results presented in this paper is  supported by the European Union's Seventh Framework Programme (FP7) under grant agreement no. 615688 (PRIME), The Israeli Science Foundation (grant number 1555/15), The Allen Institute for Artificial Intelligence, and The Intel Collaborative Research Institute for Computational Intelligence (ICRI-CI)

\bibliography{bib}
\bibliographystyle{icml2018}

\end{document}